\documentclass[a4paper, 10pt, conference]{ieeeconf}      

\IEEEoverridecommandlockouts                              

\usepackage[T1]{fontenc}
\usepackage[utf8x]{inputenc}
\usepackage[pdftex]{graphicx}
\usepackage{float}
\usepackage{amsmath}
\usepackage{bm}
\usepackage{url}

\title{\LARGE \bf
Genetic Algorithm to Optimize Design of Micro-Surgical Scissors}

\author{Fatemeh Norouziani$^{1}$, Veerash Palanichamy$^{2}$, Shivam Gupta$^{3}$, and Onaizah Onaizah$^{4}$
\thanks{$^{1,2,3,4}$The authors are with the Department of Computing \& Software, McMaster University, Ontario, Canada (emails: {\tt\small norouf1@mcmaster.ca, palanicv@mcmaster.ca, gupts44@mcmaster.ca, onaizaho@mcmaster.ca})}%
}

\begin{document}

\maketitle
\thispagestyle{empty}
\pagestyle{empty}

\begin{abstract}

Microrobotics is an attractive area of research as small-scale robots have the potential to improve the precision and dexterity offered by minimally invasive surgeries. One example of such a tool is a pair of micro-surgical scissors that was developed for cutting of tumors or cancerous tissues present deep inside the body such as in the brain. This task is often deemed difficult or impossible with conventional robotic tools due to their size and dexterity. The scissors are designed with two magnets placed a specific distance apart to maximize deflection and generate cutting forces. However, remote actuation and size requirements of the micro-surgical scissors limits the force that can be generated to puncture the tissue. To address the limitation of small output forces, we use an evolutionary algorithm to further optimize the performance of the scissors. In this study, the design of the previously developed untethered micro-surgical scissors has been modified and their performance is enhanced by determining the optimal position of the magnets as well as the direction of each magnetic moment. The developed algorithm is successfully applied to a 4-magnet configuration which results in increased net torque. This improvement in net torque is directly translated into higher cutting forces. The new configuration generates a cutting force of 58 mN from 80 generations of the evolutionary algorithm which is a 1.65 times improvement from the original design. Furthermore, the developed algorithm has the advantage that it can be deployed with minor modifications to other microrobotic tools and systems, opening up new possibilities for various medical procedures and applications.

\end{abstract}

\section{INTRODUCTION}

Minimally invasive surgery, often referred to as keyhole surgery, has become increasingly prevalent in contemporary medical practices due to the numerous advantages it offers \cite{Bao2022}. It reduces trauma, minimizes scarring, reduces risks associated with infection, and accelerates recovery time \cite{Nelson2010}. The robotic instruments currently employed in minimally invasive surgical procedures lack flexibility across various applications and are not designed to perform effectively through small incisions \cite{Onaizah2019}. The last decade has seen a lot of advances in the field of small-scale robotics with sizes ranging from micrometers to millimeters. These robots are useful as they can easily travel through small incisions or can be deployed through injections and perform a variety of operations such as targeted drug delivery \cite{Dogangil2008}, brachytherapy \cite{Devlin2007}, and biopsy \cite{Hong2022}.

Magnetically actuated tools hold significant promise for executing challenging surgeries through small incisions, as they can be controlled wirelessly from outside the body, offering high levels of dexterity and precision. One such tool reported in the literature is an untethered pair of magnetically actuated scissors \cite{Onaizah2019}. These scissors are composed of titanium sheets, a nitinol wire that acts as a restoring spring, and two NdFeB (Neodymium Iron Boron) Magnets. The size of the scissors is 15 mm by 15 mm. The maximum cutting force generated by scissors is proportional to the external magnetic flux density applied. At 20 mT, the scissors can generate approximately 35 mN from a single blade. These millimeter-size scissors have great potential for in-vivo tissue cutting and conducting biopsies. However, the size of the scissors is currently not suitable to carry out in-vivo procedures  \cite{Onaizah2019}. Moreover, the cutting force and torques may not be large enough to cut through tissues. One of the reasons is the restriction on the size of magnets for producing large forces. Comparatively large magnets are required to have higher magnetic moments as this parameter scales with the volume of magnetic material. Another reason could be attributed to the friction present when the scissors have to be closed and blades come in contact with each other. This can significantly lower the force required for cutting, as a portion of the energy is otherwise expended in overcoming friction \cite{Nelson2010,Enomoto2021}. It is important to model the forces used for cutting biological tissues \cite{Mahvash2008,Yang2010} as well as determine the position of magnets for optimizing the cutting forces.

There are various studies in the literature where evolutionary algorithms are used for optimizing robot design. In \cite{Kim2022}, Kim et al. used a genetic algorithm to optimize design parameters for a transoral robotic system, while in \cite{Fei2023}, Fei et al. used it to design a modular robot topology based on the specified task. An evolutionary algorithm is to emulate biological evolution in which multiple different solutions are evaluated against each other over multiple iterations and the fittest solutions are kept over time. The set of solutions is also modified through mutations and crossover between individuals, eliminating some of the poorly performing solutions. The two main components of implementing a genetic algorithm are the genome and the fitness function. The genome is the encoding for each individual solution to our optimization problem, and the fitness function determines the performance of our solution. The use of genetic evolution for design optimization allows us to design an algorithm that is adaptable for different problem scopes and parameters. The algorithm designed in this paper can easily be modified to work with various other types of microrobotic tools.

\begin{figure}[]
    \centering
    \includegraphics[width=1\linewidth]{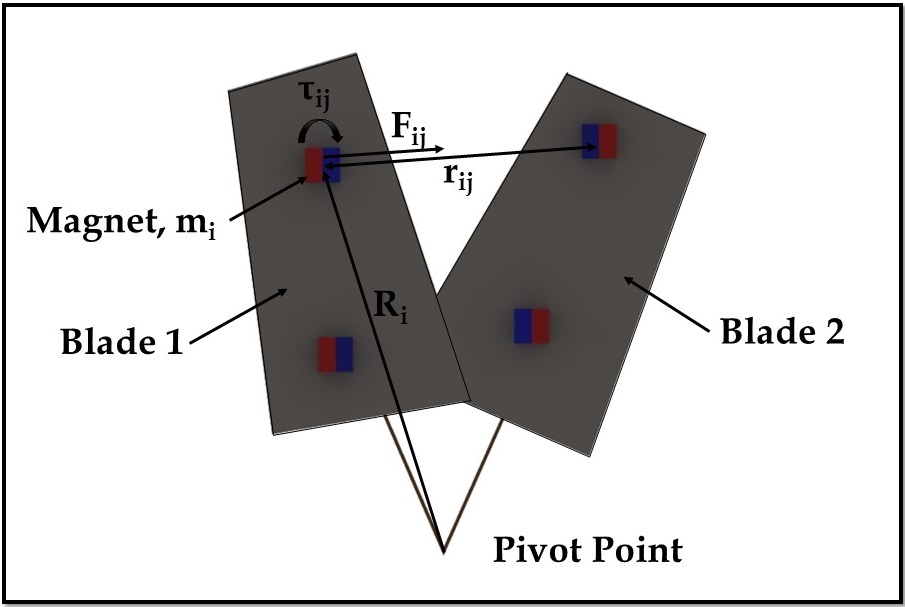}
    \caption{Schematic labeling the radii as well as interaction forces and torques between the magnets.}
    \label{fig:scissor}
\end{figure}

In this study, we introduce an evolutionary algorithm to tackle the problem of optimizing magnet position and orientation to generate a higher cutting force. Specifically, we present a 4-magnet configuration instead of the original 2-magnet design to increase the total magnetic moment available on the scissor while keeping the magnets the same size. Section II lays out the underlying physical principles and modeling as well as algorithm design. Section III discusses the results of the design optimization and presents the final configuration. Section IV presents the conclusions and future directions for the developed algorithm.

\section{Methods}
The physical model of the scissors along with the algorithm design and fitness function are presented in this section.
\subsection{Background}
The physical principles of the scissors are modeled as a torque balance. The force between two dipoles $i$ and $j$ having magnetic moment $\bm{m_i}$ and $\bm{m_j}$ can be modelled as:\\

\begin{multline} 
\bm{F_{ij}}=\frac{3 \mu_0}{4\pi r^5} [\bm{( m_i \cdot r_{ij}) m_j + (m_j \cdot r_{ij})m_i}\\ + \bm{(m_i \cdot m_j)r_{ij}} - \frac{5(\bm{m_i \cdot r_{ij}})(\bm{m_j \cdot r_{ij}})}{r^2} \bm{r_{ij}}]\\  
\end{multline}   
where $\bm{r_{ij}}$ is a vector from $\bm{m_i}$ to $\bm{m_j}$ and $\bm{r_{ij}}$ is the magnitude of $\bm{r_{ij}}$. Figure \ref{fig:scissor} shows a schematic of the interaction forces and torques between the magnets.

If $\bm{R_{i}}$ is a vector from the pivot point to any dipole $i$, then the resulting torque from the interaction force given above in Equation 1 can be calculated as:

\begin{equation}
 \bm{\tau_{fi}}= \bm{R_i \times F_{ij}}
\end{equation}

\noindent The torque generated by dipole $\bm{m_i}$ on $\bm{m_j}$ is given by:
\begin{equation}
 \bm{\tau_{ij}}= \frac{\mu_0}{4\pi r^5} [3\bm{m_j \times (m_i \cdot r_{ij})r_{ij}} - r^2 \bm{(m_j \times m_i)}]  
\end{equation}
 When no external field is applied, the scissors are in an equilibrium which can be represented as:
\begin{equation}
  \bm{\tau_{fi} + \tau_{ij}} - k \bm{p_{0}} = 0 
\end{equation}
Here, $k$ denotes the spring constant of the nitinol wire, $\mathbf{p_{0}}$ denotes the initial deflection of the spring when no external field is applied.\\

If an external magnetic flux density $\bm{B}$ is applied then torque on a magnet with magnetic moment $\bm{m_i}$ is given by:\\
\begin{equation}
    \bm{\tau_{m_i}} = \bm{m_i \times B}
\end{equation}

Therefore, the total torque on the scissors' blade is given by:

\begin{equation}
     \bm{\tau_{net}} = \sum_{i,j=1}^{n} \bm{(\tau_{fi} + \tau_{ij} + \tau_{m_i})}
     \label{eq:totalt}
\end{equation}

Here, $i,j$ represents th $n$th magnet where $n$ is taken as 4 for this study. Moreover, i $\neq$ j and any components of $\bm{\tau_{ij}}$ in the equation above should be discarded for magnets on the same blade. For example, if magnets 1 and 2 are on the same blade, the $\bm{\tau_{12}}$ is discarded from the above equation as magnets on the same blade have no relative motion and will not contribute to net torque on the blade.

\subsection{Representation}
The parameters we are trying to optimize are the magnet position and the orientation of each magnet which can be represented by a $3n$ genetic vector of the shape:
\begin{equation*}
    <x_1, y_1, M_{x_1}, ..., x_n, y_n, \theta_n>
\end{equation*}
where $x_i, y_i$ represent the x and y position of the ith magnet while $\theta_i$ represents the orientation of the magnet from the positive x-axis. Note that the magnetization vector $\bm{m}$ is calculated from the angle $\theta_i$ using the trigonometric relationships which are then used in the equations.

\subsection{Evaluation Function}
In this case, the net torque is taken as the fitness function that will be maximized and is shown in Equation \ref{eq:totalt}. A penalty of value $p$ is subtracted from the fitness score if an individual solution represents an invalid configuration. Two ways for an invalid configuration to occur are when either one or more of the magnets are outside the allowed region (blade geometry) or if they are placed less than one magnet distance away from each other. 

Since we want the scissors to be able to open to the same separation level as the original design when a magnetic field is not present, we will have to make sure the magnetic interaction torque ($\bm{\tau_{fi} + \tau_{ij}}$) is less than the threshold torque that would result in having a smaller separation between the scissor blades. This check is also done during fitness evaluation and given a penalty if violated.

\subsection{Evolution}
The initial population is generated randomly using a uniform distribution while making sure they are within the boundaries of the scissors and do not have magnets that are too close.
The population for the next generation is based on the previous generation. The population evolves solely based on the initial individuals and the genetic operations applied to them. Within each generation, individuals undergo various mutations and crossover, and it is vital to ensure that the modified genes meet the restrictions mentioned above that are controlled through the fitness function.

Specifically, we use an Evolutionary Algorithm (EA) with a ($\mu,\lambda$) selection mechanism from the DEAP library (Python 3.11) to update the population in each generation. $\mu$ represents the number of parents selected, and $\lambda$ represents the number of offspring that form the next generation. In the ($\mu,\lambda$) selection strategy, the $\mu$ best offspring are chosen based on their fitness and form the next generation. The parents are not part of the next generation, ensuring that it can avoid getting trapped in a local minima \cite{Eiben2015}. The evolution parameters are specified in Table \ref{tab:evolutionary-parameters} and worked well for this problem. The specific constants used for the torque calculations that determine the fitness are given in Table \ref{tab:constants}.

\begin{table}[h]
    \centering
     \caption{Evolution Parameters in the Code}
    \begin{tabular}{lc}
        \hline
        \textbf{Parameter} & \textbf{Value} \\
        \hline
        Population Size & 70 \\
        Number of Generations & 50 \\
        Crossover Probability & 0.7 \\
        Mutation Probability & 0.2 \\
        Differential Evolution Alpha Parameter  & 0.5 \\
        Mutation Mean ($\mu$) & 0 \\
        Mutation Standard Deviation ($\sigma$) & 1 \\
        Selection Operator Tournament Size & 3 \\
        \hline
    \end{tabular}
    \label{tab:evolutionary-parameters}
\end{table}
\begin{table}[h]
    \centering
    \caption{Constants used for fitness calculations}
    \begin{tabular}{lc}
        \hline
        \textbf{Constant} & \textbf{Value} \\
        \hline
        Magnetic Moment & 3.34E-2 $kA\cdot m^2$ \\
        Magnet Length (cube) & 31.8 $mm$ \\
        Applied Field & 20 $mT$ \\
        Minimum Magnet Separation Distance & 4.77 $mm$ \\
        \hline
    \end{tabular}
    \label{tab:constants}
\end{table}

\section{Results}

\begin{figure}[]
    \centering
    \includegraphics[width=\linewidth]{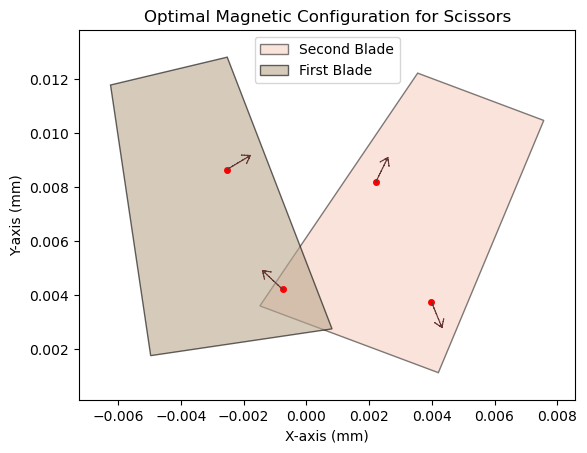}
    \caption{Figure showing the optimal configuration of magnets found using genetic evolution. The red dots represent the center of the magnets while the arrows represent the magnetization direction}
    \label{fig:optimal-config}
\end{figure}

\begin{figure}[]
    \centering
    \includegraphics[width=1\linewidth]{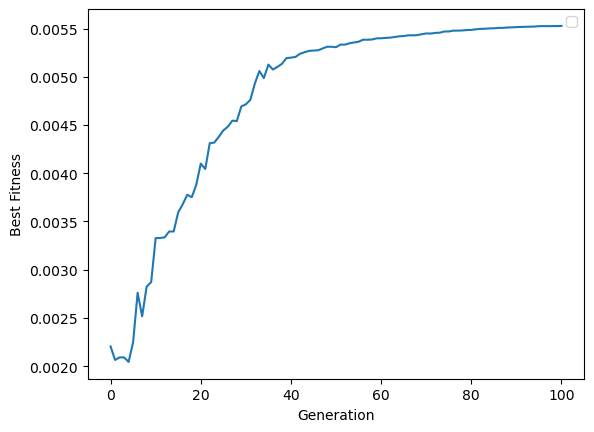}
    \caption{Graph showing the maximum fitness score throughout evolution. The fitness function starts to saturate around 80 generations beyond which only minimal improvements are produced.}
    \label{fig:fitnessgraph}
\end{figure}

The generational convergence of the algorithm is tracked using maximum fitness scores for each configuration which is shown in Figure \ref{fig:fitnessgraph}. Most of the evolution happens early, before 40 generations, where it follows a linear progression. The algorithm seems to converge around 80 generations to an optimal fitness of 5E-3 as shown in Figure \ref{fig:fitnessgraph}. After this convergence, further training is less likely to improve our fitness significantly. The maximum fitness of 5.5E-3 represents a maximum torque $\bm{\tau_{max}}$ of 5.5 mN-m for a single blade. Since this total torque does not include the restoring torque from the nitinol wire in our fitness calculation, the resulting torque can then be calculated using:
\begin{equation}
    \bm{\tau_{final}} = \bm{\tau_{max}} + \bm{\tau_{spring}}
    \label{eq:final}
\end{equation}
where $\bm{\tau_{spring}} = k \bm{p_{0}}$, and $k$ is the spring constant that can be calculated from the specifications of the nitinol wire. Solving Equation \ref{eq:final} gives us a torque of 4.6 mN-m, from which the force is calculated for a single blade. The resulting cutting force is 58 mN which is 1.65 times higher than the cutting force that was measured from the original design to be 35 mN for a single blade.

The optimal configuration is shown in Figure \ref{fig:optimal-config}. The algorithm prioritizes putting the magnets as close to each other while satisfying the distance constraint of not being within one magnet length of each other. Another interesting assessment is that the magnet on the bottom left has a magnetization vector with its y-component being negative. Although this orientation may not allow for the maximum torque when a field is applied along the positive y-axis, this orientation is necessary for the magnet configuration to balance the magnetic interactions that could lead to the scissors having more than our desired equilibrium deflection.  Although we are assuming magnets of the same size for our theoretical torque calculations and in our evolutionary algorithm, smaller magnet sizes can be used in this optimal configuration to generate the same cutting force as the original design. This will help reduce the size of the microrobot and the weight of the proposed design and prototype.

\section{Conclusion}
In this paper, we have shown that it is possible to use genetic evolution to create an adaptable algorithm that can help generate design parameters for microrobots. We specifically applied our methodology to generate magnet position and orientation to improve the design of our previously developed minimally invasive micro-surgical scissors. The new proposed configuration has 4 magnets compared to the 2 magnets seen in the original. This new design can generate approximately 1.65 the cutting force of the original scissors at 58 mN. Our optimal configuration also allows the option for the device to be scaled down while keeping similar cutting forces of 35 mN that are seen in the original design which is an important and desirable characteristic of microrobot design.

Future work will be done to redefine this problem using a multi-objective evolutionary algorithm. Such an algorithm might be more suitable because the scissors need to be able to open by themselves when a magnetic field is not present in order to use only control input. Magnet configurations that improve the maximum cutting torque negatively affect the restoring torque of the scissors. This demands a 2-objective maximization problem which can provide a more diverse solution set. Future research will also be done to extend our algorithm to other types of devices \cite{Lim2021DesignNeuroendoscopy,Forbrigger2019Cable-LessSurgery} that may need similar design optimization.





\bibliographystyle{IEEEtran}
\bibliography{refs, references}

%

\end{document}